\documentclass{article}

\usepackage{mathtools}
\usepackage[margin=1.0in]{geometry}
\usepackage{parskip}
\usepackage{fancyhdr,graphicx,amsmath,amssymb}
\usepackage[ruled,vlined]{algorithm2e}
\usepackage{bbm}
\usepackage{mathrsfs}
\usepackage{cancel}
\usepackage{svg}
\usepackage{float}

\DeclarePairedDelimiter{\ceil}{\lceil}{\rceil}
\DeclarePairedDelimiter{\floor}{\lfloor}{\rfloor}

\usepackage{hyperref}       %
\usepackage{url}            %
\usepackage{amsfonts}       %
\usepackage{nicefrac}       %
\usepackage{microtype}      %
\usepackage{forloop}
\usepackage{xcolor}
\usepackage[hang,flushmargin]{footmisc}

\DeclareMathOperator{\KL}{\mathcal{KL}}

\newcommand{\defvec}[1]{\expandafter\newcommand\csname v#1\endcsname{{\mathbf{#1}}}}
\newcounter{ct}
\forLoop{1}{26}{ct}{
    \edef\letter{\alph{ct}}
    \expandafter\defvec\letter
}
\forLoop{1}{26}{ct}{
    \edef\letter{\Alph{ct}}
    \expandafter\defvec\letter
}

\newcommand{\field}[1]{\ensuremath{\mathbb{#1}}}

\newcommand{\reals}{\field{R}}

\DeclareMathOperator*{\argmax}{\arg\!\max}

\makeatletter
\newcommand{\algorithmfootnote}[2][\footnotesize]{%
  \let\old@algocf@finish\@algocf@finish%
  \def\@algocf@finish{\old@algocf@finish%
    \leavevmode\rlap{\begin{minipage}{\linewidth}
    #1#2
    \end{minipage}}%
  }%
}
\makeatother

\title{Non-parametric generalized linear model}

\author{%
  Matthew Dowling
  \hfill
  Yuan Zhao
  \hfill
  Il Memming Park
  \\
  Center for Neural Circuit Dynamics\\
  Stony Brook University, Stony Brook, NY 11794-5230, USA
  \\
  \texttt{\{matthew.dowling,yuan.zhao,memming.park\}@stonybrook.edu}
}

\begin{document}

\maketitle

\begin{abstract}

  A fundamental problem in statistical neuroscience is to model how neurons encode information by analyzing electrophysiological recordings.
  A popular and widely-used approach is to fit the spike trains with an autoregressive point process model.
  These models are characterized by a set of convolutional temporal filters, whose subsequent analysis can help reveal how neurons encode stimuli, interact with each other, and process information.
  In practice a sufficiently rich but small ensemble of temporal basis functions needs to be chosen to parameterize the filters.
  However, obtaining a satisfactory fit often requires burdensome model selection and fine tuning the form of the basis functions and their temporal span.
  In this paper we propose a nonparametric approach for jointly inferring the filters and hyperparameters using the Gaussian process framework.
  Our method is computationally efficient taking advantage of the sparse variational approximation while being flexible and rich enough to characterize arbitrary filters in continuous time lag.
  Moreover, our method automatically learns the temporal span of the filter.
  For the particular application in neuroscience, we designed priors for stimulus and history filters useful for the spike trains.
  We compare and validate our method on simulated and real neural spike train data.
\end{abstract}

\section{Introduction}
In neuroscience, statistical modeling of the influence of external covariates and self-history to neuronal activity provides an avenue to study the neural code used by the neural system for processing sensory stimuli, cognitive computation, and motor control.
For neurons where the observations are sequence of all-or-none action potentials, or spike trains, the field has widely adapted the use of an autoregressive point process model, often referred to as the generalized linear model (GLM) or the nonlinear Hawkes process~\cite{Truccolo2005,pillow_spatio-temporal_2008,Truccolo2016,Hart2020}.

\begin{figure}[t!b]
    \centering
    \includegraphics[width=1.0\textwidth]{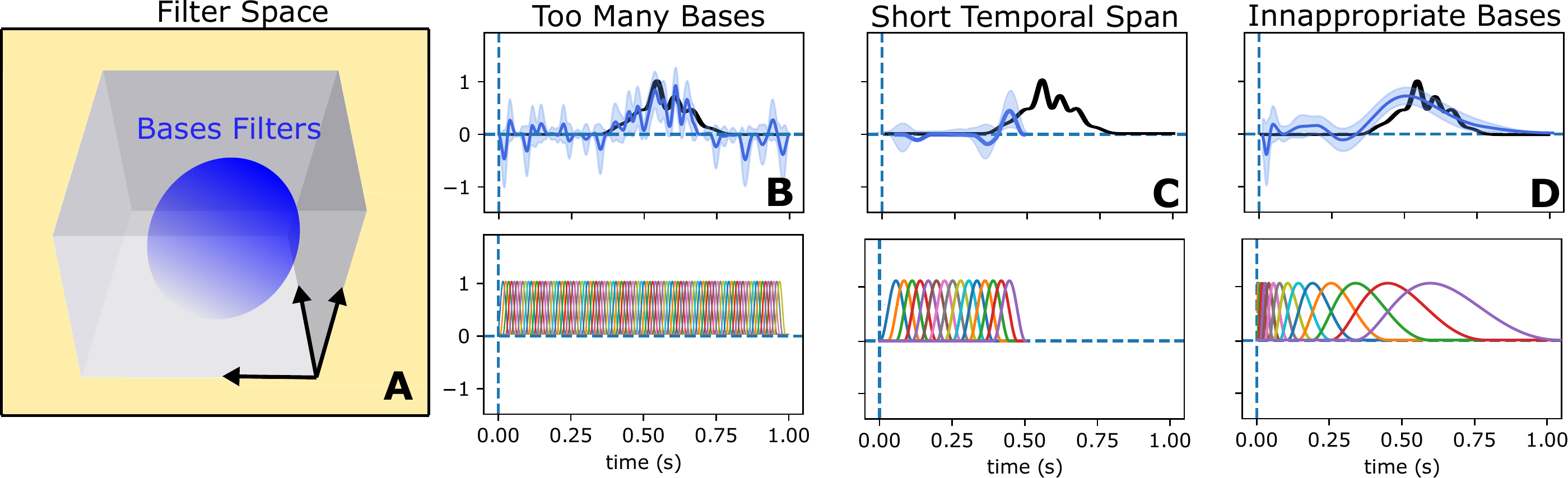}
    \caption{
        Mishaps that can occur using a fixed set of bases. \textbf{A)} For a fixed set of bases only a subset of possible filters are actually realizable.  \textbf{B)} Too many bases used resulting in difficulty capturing smooth properties seen in the ground truth.  \textbf{C)} The temporal span was under-specified and so the resulting inference is not able to capture the full extent of the filter.  \textbf{D)} A linear combination of an inappropriate set of bases has trouble inferring characteristics outside its realm. Shown below each example is the set of bases used.}
    \label{fig:filter_space}
\end{figure}

To incorporate the time-varying modulation controlled by the covariates, GLMs and GLM-like autoregressive point process models are parametrized with (time) convolutional filters~\cite{tripathy_intermediate_2013,latimer_inferring_2019}.
Given time-varying covariates $x_i(t)$, the neural spiking activity $y(t)$ is fully specified as a point process with the conditional intensity function~\cite{Truccolo2005}:
\begin{equation}
    \lambda(t;\mathscr{H}_t) = G
        \left(
            \sum_i (f_i \ast x_i)(t) + d
        \right)  %
    \label{eqn:conditional_intensity}
\end{equation}
where $f \ast x = \int f(\tau) x(t-\tau) \mathrm{d}\tau$ denotes convolution, $f_i(\tau): \reals \to \reals$ are temporal filters, $d \in \reals$ is the bias term that defines a baseline firing rate, $G(\cdot): \reals \to [0, \infty)$ is a pointwise nonlinearity, and $\mathscr{H}_t$ denotes the filtration on the past~\cite{Daley1988}.
The conditional intensity function $\lambda(t;\mathscr{H}_t)$ describes the instantaneous firing rate at time $t$ given its history.
In practice, the filters are further parameterized by a finite set of parameters:
\begin{align}\label{eq:GLMwBasis}
    f_i(\tau) = \sum_{j=1}^{N_b} \beta_{i,j} b_j(\tau)
\end{align}
where $N_b$ number of fixed basis functions $b_j: \reals \to \reals$ are weighted by coefficients $\beta_{i,j} \in \reals$.
This allows (1) efficient model fitting, (2) smoothness of $f_i$ controlled by the smoothness of $\{b_j\}$, and (3) temporal support of $f_i$ to be prespecified.
For example, in neural data analysis, the raised cosine basis set is used for covariates and its temporally log-scaled version in combination with a boxcar basis, to capture fast transition to refractory period, is used for spiking history.~\cite{pillow_spatio-temporal_2008}.
In practice, the span of the filter can last for a few hundred milliseconds in the early sensory neurons to a few seconds in higher order cortical areas.
However, the burden for the neurostatistician is to tune the \textit{hyperparameters}, e.g., the smoothness, the temporal span, and the number of basis functions.
This creates a large landscape over the space of possible models and requires extensive model selection to choose the best performing one.
These hyperparameter choices are among the top frequently asked question by the new users of GLM framework for neural data analysis, and improper choice can lead to misleading scientific conclusions or poor inference as seen in Fig.~\ref{fig:filter_space}

Here, we propose an automated method that alleviates the GLM user from manual tuning of parameters. We make use of the Gaussian Process (GP) framework to impose flexible and expressive priors over the set of filters specified in the model. Being nonparametric in nature this approach requires little prior parameterization and the ability to update hyperparameters during optimization endows us with a rich space of filters that can be captured.

\section{Method}
\subsection{Gaussian processes and sparse approximation}
A Gaussian process is a time continuous stochastic process of which every finite collection of random variables follows a multivariate Gaussian distribution~\cite{Rasmussen2005}.
GPs are widely-used for modeling functions with a probabilistic view that quantifies uncertainty and suits Bayesian inference.
We model each temporal filter $f(\tau)$ over time lag $\tau \in \mathfrak{T} \subset \mathbb{R}$ with a GP:
\begin{equation}
    f(\tau) \sim \mathcal{GP}(m(\tau), k(\tau, \tau'))
\end{equation}
which is fully specified by the mean function $m(\tau)$ and covariance function $k(\tau, \tau')$ (a.k.a. kernel).

For a finite set of points (lag times) $\boldsymbol{\tau} = \{\tau_n\}_1^N$, the corresponding values $\vf = \{f(\tau_n)\}_1^N$ are Gaussian distributed s.t.
\begin{equation}
    \vf \sim \mathcal{N}(\vm_{\boldsymbol{\tau}}, \vK_{\boldsymbol{\tau}})
\end{equation}
where $\vm_{\boldsymbol{\tau}} = [m(\tau_n)]_{n=1,\ldots,N}^\top$ and $\vK_{\boldsymbol{\tau}} = [k(\tau, \tau')]_{\tau, \tau' \in \boldsymbol{\tau}}$.
One can interpolate or extrapolate the value of the filter at any new time $\tau^*$ by
\begin{equation}
    p(f^* \mid \vf) = \mathcal{N}(m(\tau^*) + \vk_{*\boldsymbol{\tau}}^\top \vK_{\boldsymbol{\tau}}^{-1}\vm_{\boldsymbol{\tau}}, k_{*} - \vk_{*\boldsymbol{\tau}}^\top \vK_{\boldsymbol{\tau}}^{-1}  \vk_{*\boldsymbol{\tau}})
\end{equation}
where $f^* \triangleq f(\tau^*)$, $\vk_{*\boldsymbol{\tau}} = [k(\tau^*, \tau)]_{\tau \in \boldsymbol{\tau}}^\top$ and $k_{*} = k(\tau^*, \tau^*)$. 

Though Gaussian processes have nice properties, the $\mathcal{O}(N^3)$ computational complexity of training obstructs scalable and efficient application to even moderately sized problems. 
To alleviate this obstacle, we use the sparse approximation to Gaussian processes~\cite{snelson2006sparse, titsias2009variational}.
The sparse GP approach introduces a set of inducing variables $\vu = \{f(z_m)\}_{m=1}^M$ that are the filter evaluated at so-called \textit{inducing points} $\{z_m\}_{m=1}^M \subset \mathfrak{T}$. 
Moreover these inducing variables are assumed to be a sufficient statistic for $f^*$ s.t.
\begin{equation}
    p(f^* \mid \vf, \vu) =  p(f^* \mid \vu).
\end{equation}
The sparse approximation provides a finite small set of support and reduces the complexity of posterior inference to $\mathcal{O}(N M^2)$ where $M \ll N$. 

\subsection{Discretization}

With the conditional intensity as defined in \eqref{eqn:conditional_intensity}, given the spike times $\{t_n\}_1^N$, the point process log-likelihood of the interval $[0,T]$ is
\begin{equation}
    \log p(\{t_n\} \mid \{x_i(t)\}, \{f_i(\tau)\}, \theta) = \sum_n\log\lambda(t_n) - \int_0^T \! \lambda(t) \,\mathrm{d} t    \label{eqn:pp_ll}
\end{equation}
where $\theta$ represents the parameters~\cite{Daley1988}.
In order to evaluate~\eqref{eqn:pp_ll}, we have to calculate the integral $\int_0^T \! \lambda(t) \,\mathrm{d} t$. This is however often computationally costly. Moreover, many covariates such as stimuli are given or can be discretized piecewise constant along with the filter.
Dividing the interval $[0,T]$ into $K$ evenly spaced bins of width $\Delta$, and defining $\lambda_k = \lambda(k\Delta)$ for $k=0,\ldots, K-1$ the log-likelihood is commonly approximated as~\cite{Truccolo2005},
\begin{equation}
    \log p(\{t_n\} \mid \{x_i(t)\}, \{f_i(\tau)\}, \theta) = \sum_n\log\lambda(t_n) - \sum_k \Delta\lambda_k
\end{equation}

Meanwhile the discretization also involves discrete convolution of which requires the value of filter to be evaluated for each time bin.
Limited by finite amount of data and computational resources, the filters need to be truncated into a finite window $[\tau^{(L)}_i, \tau^{(U)}_i]$ at the tails. 
Supposing the evaluation of the $i$-th filter to be $\tilde{\vf}_i = \{f_i(\tau)\}_{\tau \in \{\tau^{(L)}_i, \tau^{(L)}_i + \Delta, \ldots, \tau^{(U)}_i\}}$. Then the discrete convolutions can be written as matrix vector multiplications s.t. $\vX_i \tilde{\vf}_i$ where $\vX_i$ is the design matrix $\vX_{i,nj} = x_i(n\Delta - (\tau_i^{(L)} + j\Delta))$. The log-likelihood can then be written compactly as
\begin{equation}
    \log p(\vy \mid \{\vX_i\}, \{\tilde{\vf}_i\}, \theta) = \vy^T (\sum_i \vX_i \tilde{\vf}_i + \mathbbm{1} d) - \Delta \mathbbm{1}^T \exp(\sum_i \vX_i \tilde{\vf}_i + \mathbbm{1}d)
\end{equation}
where $\mathbbm{1}$ is the all-ones vector and $\vy \in \mathbb{R^K}$ is the binary vector of discretized spikes.

\subsection{GP mean and covariance kernel selection}
The kernel of the GP prior has a strong influence on the characteristics of the posterior distribution.
Biophysical characteristics of the neural system is such that the filters for external covariates are (1) local in time lag, (2) temporally smooth, and (3) have fading memory and typically decay to zero exponentially fast.
The decaying squared exponential (DSE) kernel~\cite{tobar_learning_2015} fulfills these conditions. Moreover, to capture filters that are not centered around zero we introduce an offset $\beta$ to the DSE kernel as follows:
\begin{equation}
    k_{\text{DSE}}(\tau, \tau') = \sigma^2 e^{-\alpha(\tau - \beta)^2 -\nu(\tau - \tau')^2 -\alpha(\tau' - \beta)^2}
\end{equation}
where $\alpha$ controls the exponential decay rate, $\nu$ controls the spectral content, and $\sigma^2$ controls the finite power.

In many cases a null mean is used to specify the GP prior as this does not constrain the posterior mean.  However, it can be beneficial to incorporate prior knowledge and bias the model toward an asymptotic regime by introducing a nontrivial parameterization of the mean.  For example, the refractory period exhibited by neurons makes it very unlikely that spikes should occur within close proximity of each other.  This manifests itself in a history filter as an exponential climb from extremely negative values to zero over a short time succeeding a spike. ~\cite{10.1162/neco_a_01021}. Since this is well known we can incorporate our neuroscientific knowledge and parameterize spike history filters as such

\begin{equation}
    f(\tau) \sim \mathcal{GP}(a e^{-\frac{\tau}{b}}, k_{\text{DSE}}).
\end{equation}
which allows for the resulting inference to more easily capture the characteristic refractory period.  Indeed, this serves a similar function to the use of a 'boxcar' basis under the alternative approach.  Furthermore, without this specification the resulting inference may gravitate towards a small length scale parameter to adequately capture any transient sharp rise and thus impede inference about smooth long timescale varying characteristics. 

\subsection{Variational inference}
In Bayesian inference all of the uncertainty over a filter $f(\tau)$ once the data $\vy$ has been observed is embedded into its' posterior distribution
\begin{equation}
    p(f(\tau) \mid \vy) = \frac{p(\vy \mid f(\tau)) p(f(\tau))}{p(\vy)}
\end{equation}

In GLMs, unfortunately, the evidence, $p(\vy)$, and thus the posterior $p(f(\tau) \mid \vy)$ are intractable for GP priors. Sampling methods are able to tackle the intractability. However they require intensive computation and do not produce analytical solutions. 
Therefore we seek approximate solutions. 
More recent works have used sparse variational approaches which take care of both the intractable posterior as well as the computational complexity of full GP~\cite{lloyd_variational_2015-1, hensman_mcmc_2015}.  Variational inference~\cite{Bishop_PRML} usually assumes a parametric form of an approximation to the posterior distribution and cast the inference problem to optimization.

Continuing with our formulation where $\vu$ are the inducing variables of the filter $f(\tau)$, a lower bound on the log marginal likelihood $\log p(\vy)$ can be derived~\cite{lloyd_variational_2015-1, dezfouli_scalable_2015} yielding the variational objective
\begin{equation}\label{eq:elbo_u}
    \mathcal{L} = \mathbb{E}_{p(f(\tau) \mid \vu) q(\vu)} \log p(\mathbf{y} \mid f(\tau)) - \KL(q(\vu) \,\|\, p(\vu)) 
\end{equation}
which is often referred to as the evidence lower bound (ELBO).
The $q(\vu)$ closest in KL divergence to the true posterior can then be obtained by maximizing the ELBO.

This lower bound can be evaluated analytically if $q(\vu)$ is selected to be a parameterized Gaussian of the form $\mathcal{N}(\vu \mid \vm, \vS)$ and the joint approximate distribution $q(f(\tau), \vu)$ follows
\begin{equation}
    q(f(\tau), \vu) = p(f(\tau) \mid \vu) q(\vu),
\end{equation}

The resulting marginal $q(f(\tau)) = \int q(f(\tau), \vu) \,\mathrm{d} \vu$ is now an approximation to the true posterior and is also a Gaussian process with mean and covariance given by
\begin{align}
    \mu(\tau) &= K_u (\tau)^T K_{uu}^{-1} \vm \\
    \Sigma(\tau, \tau') &= K(\tau, \tau') - K_u(\tau)^T K_{uu}^{-1 }K_u(\tau) + K_u(\tau)^T K_{uu}^{-1}\vS K_{uu}^{-1} K_u(\tau).
\end{align}

Letting $\vu_i = \{f_i(z_m^i)\}_{m=1}^M, \, z_m^i \in \mathfrak{T}_i$ be the inducing variables of filter $f_i(\tau)$. Eq.~\eqref{eq:elbo_u} can be extended to
\begin{equation}
\begin{split}
    \mathcal{L} = \vy^T \Bigg(\sum_i \vX_i \mu_i & + \mathbbm{1}d \Bigg)
    - \Delta \mathbbm{1}^T \exp\left(d + \sum_i \vX_i \mu_i
    + \frac{1}{2}\mathrm{diag}(\vX_i \Sigma_i \vX_i^T) \right) \\
    &+ \sum_i \KL\left(q(\vu_i) \| p(\vu_i)\right)
\end{split}    
\end{equation}
where \[\mu_i = K_{X_i, X_i} K_{u_i, u_i}^{-1} m_i\] and 
\[\Sigma_i = K_{X_i,X_i} - K_{X_i,u_i}K_{u_i,u_i}^{-1}K_{u_i,X_i} + K_{X_i,u_i}K_{u_i,u_i}^{-1} S_i K_{u_i,u_i}^{-1}K_{u_i, X_i}\]
are respectively the mean and covariance of the variational approximation $q(\vu_i)$. Now, full inference can be performed as the  gradients of the ELBO w.r.t. the variational parameters $\{\vz^i, \vm_i, \vS_i\}$ and hyperparameters $\{\boldsymbol{\theta}_i\}$ can be calculated in closed form

\subsection{Optimization}
As the variational objective is differentiable w.r.t. hyperparameters, variational parameters, and model parameters, we can maximize the ELBO 
without resorting to sampling schemes or approximations. We elect to use a full rank parameterization of the covariance of each variational Gaussian, $\vS_i$, by optimizing its Cholesky factor $\vL_i$, such that $\vS_i = \vL_i \vL_i^T$ remains in the positive semi-definite cone.
While the optimization can be performed jointly we found a coordinate ascent approach more practical and consistent.  This involves partitioning the parameters of the model into two groups -- the first group $\vV_i$, containing the variational parameters, and a second group $\vH_i$, containing the kernel hyperparameters.  

In the optimization process we also incorporate updates of the convolutional design matrix for each of the covariates.  Each design matrix, $\vX_i$, and the associated set of lags over the span  $[\tau^{(L)}_i, \tau^{(U)}_i]$ are byproducts of having to choose a finite window over the lag space to realize the convolutional interaction under our model.  When the decision to restrict this window to a certain temporal region is made it is possible that the temporal extent will be underspecified.  Depending on the extent of the underspecification the model will lose the capability to explain phenomena of the firing rate that occur on long time scales or large latencies with respect to a particular covariate.  While the GP prior on the filters provides us a posterior that is well defined over the entire lag space, optimization of the model parameters i.e. hyperparameters and variational parameters, require evaluation of the expected log-likelihood which in turn is directly affected by the temporal extent of the filter used to perform the convolutions.

\begin{algorithm}[H]
\newcommand\mycommfont[1]{\footnotesize\ttfamily\textcolor{black}{#1}}
\SetCommentSty{mycommfont}
\SetKwInput{kwReturn}{Return}
\SetKwBlock{kwRepeatUntil}{Repeat Until Convergence}{end}
\SetAlgoLined
initialize parameters: \hspace{1pt} $\vV^{(0)}_k = \{\vm_k^{(0)}, \vS_k^{(0)},
\vz_k^{(0)}\}$,\hspace{2pt} $\vH^{(0)}_k = \{\boldsymbol{\theta}_k^{(0)}, d_k^{(0)}\}$, \hspace{2pt} $k=1,\ldots,K$ \\
initialize design matrices: $\{X_i\}$ \\
Set $j=1$ \\

\kwRepeatUntil{
\tcp{for each set of variational/inducing variables and hyperparameters}
    \For{$\vP^{(j-1)} \in \{\vV_k^{(j-1)}, \vH_k^{(j-1)}\}$} {
    \tcp{for each covariate}
        \For{$i=1,\ldots,K$}{
            Set $\vP^{(j)} = \underset{\vP}{\argmax} \, \mathcal{L}$
        }
    }
    \tcp{update the design matrix for each covariate}
    \For{$i=1,\ldots,K$}{
    Set $\tau_i^{(L)} = \floor{\underset{m}\min \, {z^{(j)}_{i,m}} / \Delta} \Delta$ \\
    Set $\tau_i^{(U)} = \ceil{\underset{m}\max \, {z^{(j)}_{i,m}} / \Delta} \Delta$ \\
    Update $\mathfrak{T}_i$, $X_i$
    }
    Set $j = j + 1$ 
}
\caption{NPGLM Sequential Training Algorithm}\label{algo}
\end{algorithm}

The shortcomings of having to choose this finite duration time window can be amended during the optimization procedure thanks to the inferential power granted by having used the GP framework.  In a binned time description of the model this only amounts to evaluating the variational posterior at additional time bins.  As is natural though, we would like to be greedy and use the smallest amount of evaluations we can in order to realize the convolution operation.  In that light, we propose viewing the optimized inducing points as ``guides'' to regions of high posterior density that contribute significantly to the resulting convolution.  More specifically, if the optimized inducing points for a particular covariate's filter carve out a window $[\tau^{(L)}_i, \tau^{(U)}_i]$,  then we choose to evaluate our convolutions with those as the evaluation points of the filter's posterior.  This affords us to be greedy because under-specification of this evaluation window can be amended in an iterative procedure, and more than that we avoid evaluations at points in time that have no relevance and only serve to harm the optimization procedure. Algorithm~\ref{algo} summarizes the optimization procedure.  In practice we had the most success with this style of optimization where one set of parameters ($\vV_i$ or $\vH_i$) is optimized for one covariate at a time, although we note that joint optimization of all covariates and all parameters at once worked but was less robust.

\subsection{Adaptive Design Matrix Intuition}
\begin{figure}[t!b]
    \centering
    \includegraphics[width=0.9\textwidth]{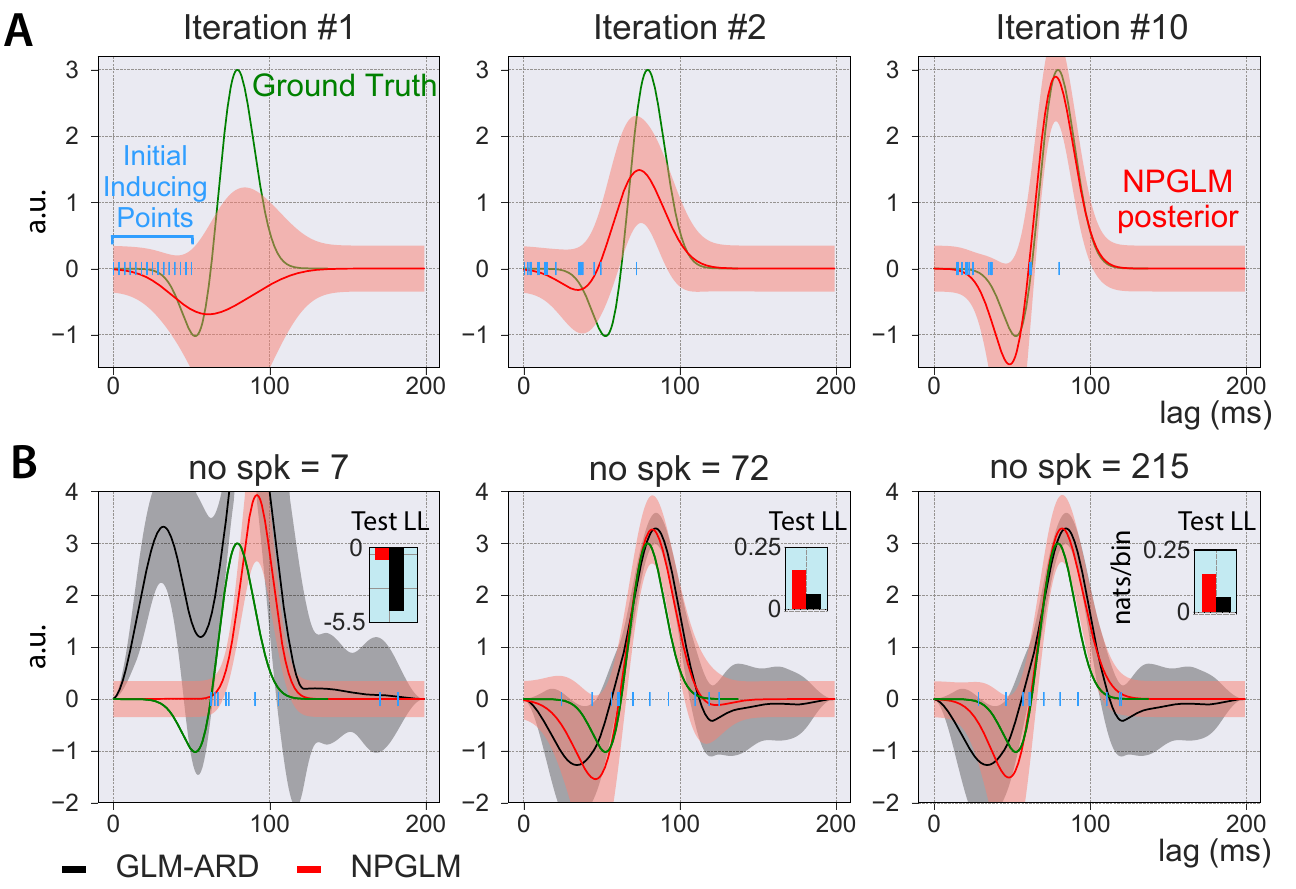}
    \caption{
        \textbf{NPGLM validation with known ground truth.}
        (\textbf{A}) 
        NPGLM recovers the span of the filter even with a bad initialization.
        NPGLM with a stimulus filter compromising 20 inducing points is initialized to have a temporal span of 50~ms.
        We plot the posterior at different optimization iterations.
        Note how not the entire span of the filter need be used when evaluating the convolutions in order to achieve a satisfactory approximation.
        Shading indicates the 95\% credible interval of the filter.
        (\textbf{B}) 
        NPGLM is robust to small sample size.
        Two Bayesian methods, NPGLM and GLM-ARD are trained on spike trains of different duration.
        Sample size from left to right: 500 (7 spikes), 5,000 (72 spikes), 15,000 (215 spikes) time bins.
        In left we see that posterior fit offered by GLM-ARD has little resemblance to the ground truth, whereas NPGLM appears more in line with the ground truth and achieves higher normalized log-likelihood.
        In middle and right the fit models both more closely resemble the ground truth with NPGLM achieving higher log-likelihood on the hold out set.
    }
    \label{fig:toy_evolution}
\end{figure}

Quickly, we would like to talk through the process of updating the design matrix, why it is necessary, and clarify further how it is done.  Take for example, inference about the underlying filter in Fig.~\ref{fig:toy_evolution}A.  Under the alternative approach of picking temporally fixed basis functions if you were to select a window of 50 $ms$ then inference about the filter would turn out poor.  Consider the approach described and an initial set of inducing points spaced uniformly from 0 $ms$ to 50 $ms$.  Now, to properly evaluate the ELBO we must convolve the discretized stimuli $x_i(n)$ with the posterior mean of the filter $\bar{\vf}$ for all $K$ time bins of observations $\vy$.  Since $\vf$ has support over all of $\mathbb{R}$ the only way to realize the convolution is to window $\vf$ which will make it 'appear' to the likelihood that $\vf$ is null outside this range. Concretely, on the first round of optimization you would have the approximation:

\begin{align*}
    \text{E}\left(\vy^T(\vx * \vf + \mathbbm{1}d) - \Delta\mathbbm{1}^T\exp(d+\vy * \vf)\right) \approx \vy^T \Bigg(\vX \mu & + \mathbbm{1}d \Bigg)
    - \Delta \mathbbm{1}^T \exp\left(d +  \vX \mu
    + \frac{1}{2}\mathrm{diag}(\vX \Sigma \vX^T) \right)
\end{align*}

\begin{figure}[t!b]
    \centering
    \includegraphics[width=1.0\textwidth]{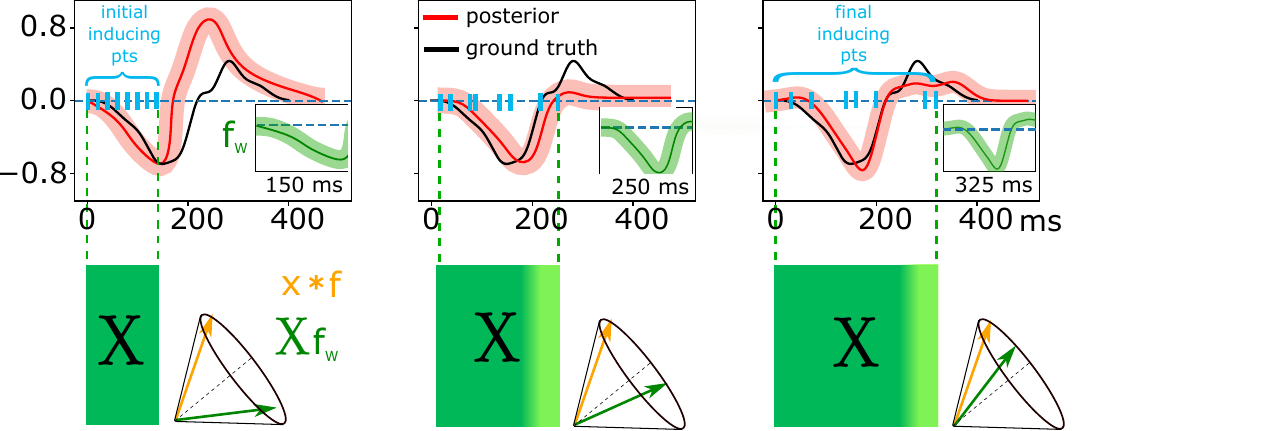}
    \caption{
        Example showing the effects of having to choose a finite window over the filter to perform convolutions.  The finite window impedes inference due to a misinformed likelihood, however, using the span dictated by the inducing points we can realize a mechanism that adaptively windows more of the relevant space.  In this way computation time can be saved by choosing a smaller initial window and allowing it to adapt over time.}
    \label{fig:inducing_intuition}
\end{figure}

where $X_i \in \mathbb{R}^{K \times 50}$ since we chose 50 $ms$ as the span of the inducing points and elected to make the cutoff based on their span.  Thus, the approximation to the likelihood will be off the mark.  What can save us is the fact that the prior will be in disagreement and so some inducing points will be 'pulled' beyond the imposed cutoff.  If this is the case the window can be increased and consequently $X$ will be expanded. Note how in Fig~\ref{fig:inducing_intuition} the range of the inducing points dictate the effective filter seen by the likelihood --- as they extend further the truncated convolution more closely matches that of one performed using the full GP posterior.

\section{Experiments}
To examine the methodology described in the paper we examine its performance on synthetic data as well as real neural recordings. In all experiments we measure performance relative to standard GLM approaches where the filters are parameterized by a set of fixed basis functions.  One GLM we compare to places an automatic relevance determination (ARD)~\cite{qi2004predictive} prior over the weights of the basis functions -- we refer to this as GLM-ARD. For the other GLM no prior is set over the weights and all inference is performed using the MLE -- we refer to this one as GLM-MLE.  The methodology described in the paper is referred to as Non-Parametric GLM (NPGLM).

\subsection{Synthetic experiment 1}
We create a stimulus filter that is a sum of two Gaussians, specifically 
\begin{equation}
    s(t)=3\,\mathcal{N}(t \mid 75, 12.5^2) - 1.8\,\mathcal{N}(t \mid 62.5, 12.5^2)
\end{equation} 
and a history filter that is supposed to mimic a short refractory period followed by a transient self-excitation.

\begin{figure}[t!b]
    \centering
    \includegraphics[width=0.9\textwidth]{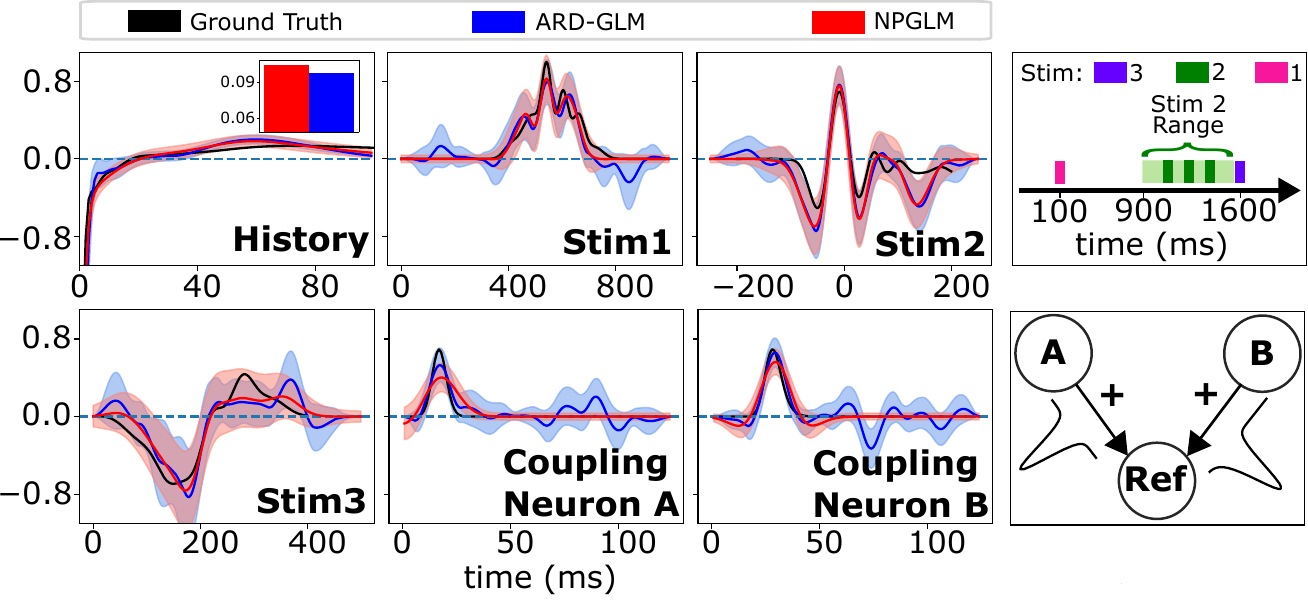}
    \caption{
        Inference over a synthetic neuron with several tuned stimuli responses, a history filter, and positive coupling from two other neurons, A and B.  NPGLM is able to accurately capture the structure of the ground truth filters.  In many cases we see the basis function approach has significant non-zero values after the ground truth has subsided to zero.  Additionally, GLM-ARD is more confident even when inference does not capture the ground truth sufficiently.  Shading represents the 95\% credible region for both approaches.  NPGLM outperforms GLM-ARD  when examing log-likelihood on a held out test set (History filter panel top right).  }
    \label{fig:synthetic_expt_2}
\end{figure}
We use this toy example to quantify A) how well the adaptive truncation window mechanism functions and B) how inference scales with the amount of relevant data. Figure~\ref{fig:toy_evolution}B shows the resulting stimulus filters after fitting NPGLM and GLM-ARD for various length spike trains.  When there are only 7 spikes in the entire sequence the posterior under GLM-ARD does not bear much resemblance to the ground truth used for generating the data.  NPGLM also has trouble capturing the ground truth but qualitatively speaking the fit appears much closer. For the two other examples where 72 and 215 spikes were used when fitting the models both capture the ground truth reasonably well. However the wider tails present in the GLM-ARD fit do end up adversely affecting its evaluation on the test set.

To illustrate the ability of NPGLM to adaptively learn the temporal span of the filter in Figure~\ref{fig:toy_evolution} we initialize the stimulus filter so that the most extreme inducing point is right on the cusp of an influential part of the filter.  Specifically 10 inducing points are initialized uniformly from 0~ms to 50~ms. As described in Algorithm~\ref{algo} migration of the inducing points towards extreme locations in space leads to adaption of the design matrix in turn leading to a larger effective window to evaluate the convolutions.  Following this strategy, it is not long before the inducing points have migrated far enough that the posterior bears a strong resemblance to the ground truth.

\subsection{Toy Experiment 2}
Now, we examine a more complicated example.  This time three stimuli and two coupling filters are used to generate the synthetic data.  Each of the stimuli are associated with a distinct neural response dictated by the appropriate filter.  Further, two neurons 'A' and 'B' are coupled positively at different latencies to our 'reference' neuron.  NPGLM is initialized with inducing points spaced 15 $ms$ apart for the stimuli filters, 3 $ms$ apart for the history filter, and as before the DSE kernel is chosen for all GPs involved.  GLM-ARD is fit using the raised cosine basis and the temporal span is chosen to adequately cover each of the filters.

\begin{figure}[H]
    \centering
    \includegraphics[width=0.8\textwidth]{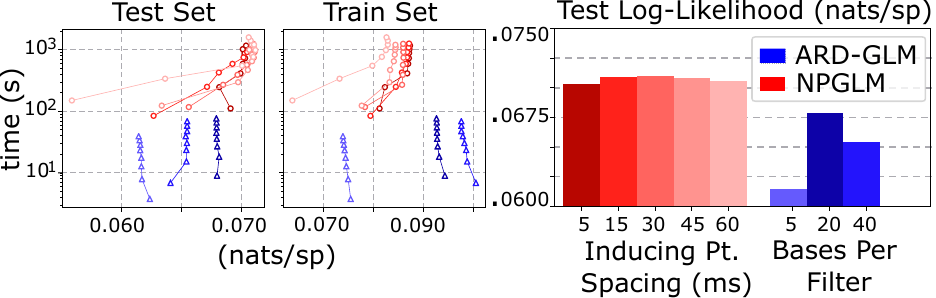}
    \caption{Run times as a function of the number of inducing points.  NPGLM is robust against the number of inducing points chosen. We can see using additional inducing points does not increase run times drastically due to the round-robin style of optimization. In contrast quality of inference when using a set of bases functions will depend significantly on their parameters.}
    \label{fig:runtimes}
\end{figure}

In Fig.~\ref{fig:synthetic_expt_2}  we can see for each of the synthetic filters that NPGLM is able to adequately capture their identifying features.  The alternative, GLM-ARD is able to as well although it appears to struggle maintaining smoothness in the tails of the coupling filters.  Additionally, we can note some spurious sharp features inferred in some of the stimuli filters where they were predominantly smooth.

We then used this same example to quantify the run time of NPGLM as a function of the number of inducing points.  For each of the stimuli filters inducing points varied from spacings of 5 $ms$ apart (more inducing points) to spacings of 60 $ms$ apart (less inducing points),  for example, in the extreme case of 60 $ms$ spacing the filter for Stim3 uses only 7 inducing points. Examining Fig.~\ref{fig:runtimes} run times are longer than GLM-ARD, however, this is to be expected and it is clear that full optimization is not required for a satisfactory level of inference.  Furthermore, even though NPGLM is not convex in it's parameters it achieves consistent, better inference, and is robust to the number of inducing points chosen.

\begin{figure}[tb]
    \includegraphics[width=\textwidth]{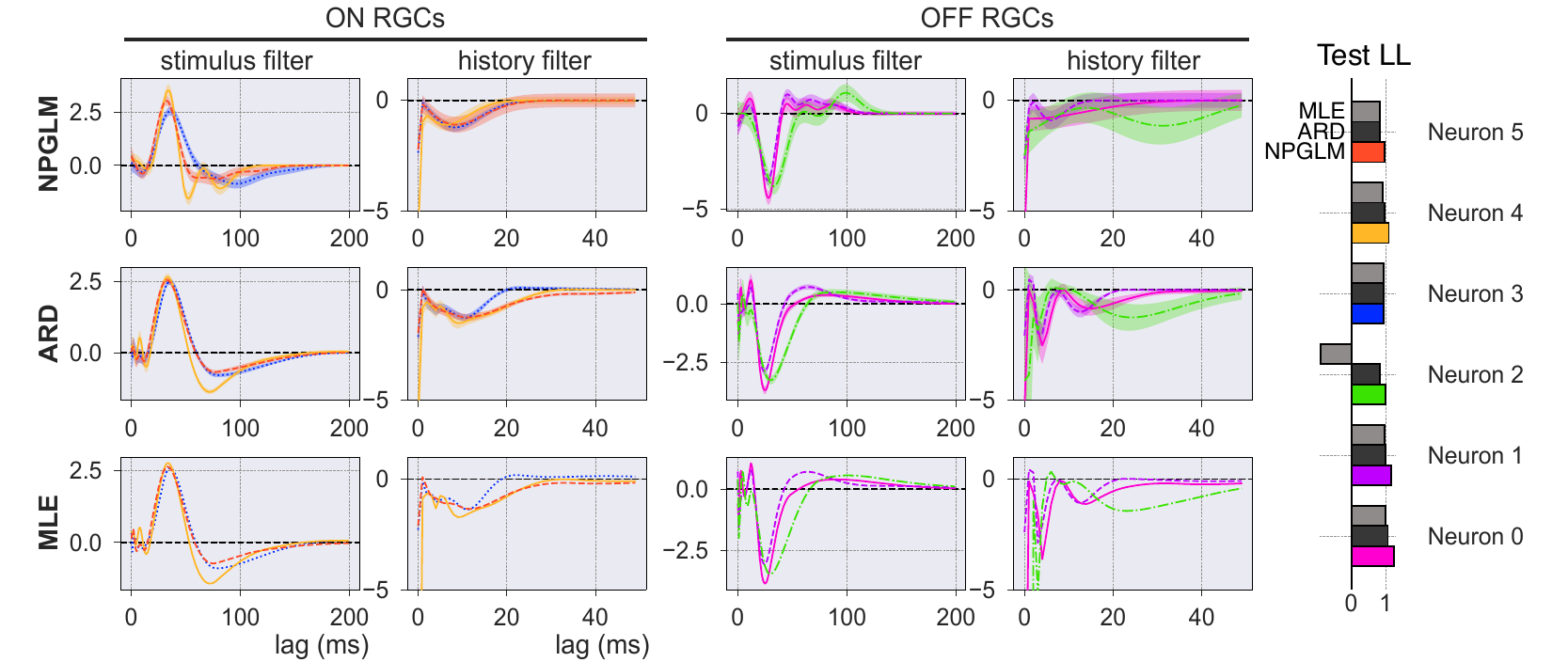}
    \caption{
        \textbf{NPGLM inference on six individual ON/OFF retinal ganglion cells is consistently superior to baseline models.}
        (\textbf{A}) Filters inferred by NPGLM using data from~\cite{Pillow2005}.
        For the most part the filters fitted to the NPGLM model strongly resemble the GLM-MLE and GLM-ARD fits.
        For certain neurons we see that different features may make themselves more well known i.e. neuron 4 and neuron 2.
        See Fig.~\ref{fig:RGC05:summary} for further details.
        (\textbf{B}) GLM-ARD fits.
        (\textbf{C}) GLM-MLE fits.
        (right) normalized test log-likelihood comparison shows that NPGLM consistently outperforms.
    }
    \label{fig:RGC05:all}
\end{figure}

\begin{figure}[tb]
    \centering
    \includegraphics[width=\textwidth]{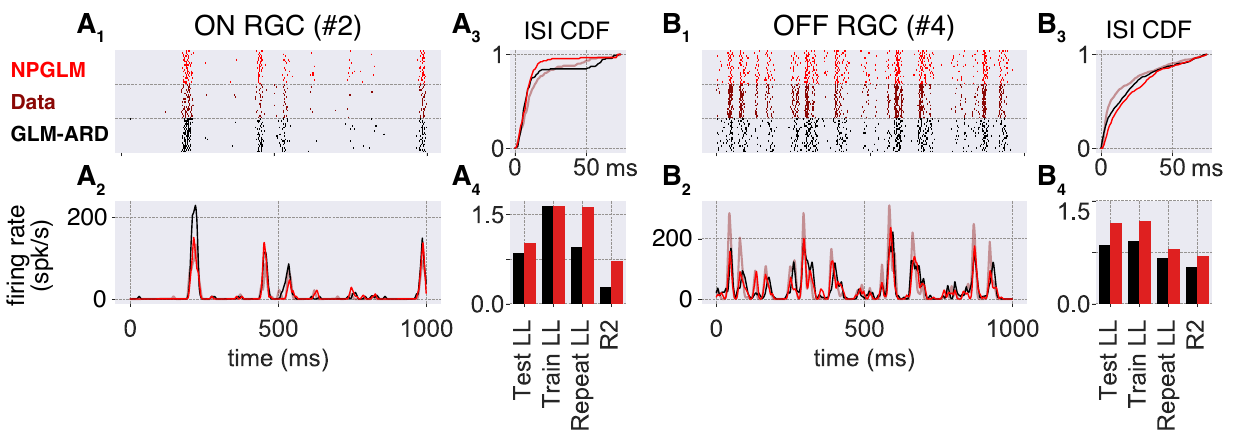}
    \caption{
        \textbf{Spike trains generated from NPGLM outperforms baseline.}
        Two example retinal ganglion cells (\textbf{A} and \textbf{B}) fit with NPGLM and GLM with basis functions (ARD w/b).
        Spike trains are simulated using ancestral sampling.
        \textbf{(1)} Raster plot of data and model simulation (top: NPGLM; middle: data; bottom: ARD). The identical random stimulus was repeated for multiple trials. 
        \textbf{(2)} Average firing rates estimated from data and simulations.
        \textbf{(3)} Empirical cumulative distribution of inter-spike intervals.
        \textbf{(4)} Goodness-of-fit measures. From left to right: normalized log-likelihood (nats/bin) for training, test and repeat datasets, and $R^2$ of firing rate for the repeat dataset.
    }
    \label{fig:RGC05:summary}
\end{figure}
\subsection{Retinal ganglion data}
We examine \textit{in vitro} multielectrode extracellular recordings from the retina of macaque monkeys~\cite{Pillow2005}. The monkeys were presented a spatially uniform cathode ray tube display refreshing at 120~Hz. Each frame took on one of two intensity values.  Previous analyses served to quantify the statistical differences in behavior between ``ON'' and ``OFF'' -- so the data set offers an established testbed for model comparison.  For this example NPGLM, GLM-ARD, and GLM-MLE are all fit to the recorded data and the bases used for both filters in the latter approaches is a series of raised cosine bumps scaled logarithmically in time.  For the stimuli filter 15 such bases spanning 300~ms are used and for the history filter 15 such basis spanning 100~ms are used.  NPGLM uses 15 inducing points spaced uniformly in time for the stimulus filter and 25 inducing points spaced logarithmically in time for the history filter.

For all six neurons the fit under each model is shown in Figure~\ref{fig:RGC05:all}.  The normalized log-likelihood for a novel repeat stimulus was also calculated to quantiatively compare how well each model performed.  For all of these cases except one we saw that the NPGLM approach performed better.  In the case of Neuron 2 it appears that NPGLM was unable to capture a sharper rise in the history filter occurring at 10~ms.  In practice the history filter is the more difficult of the two to capture because of its non-smooth behavior.

In Figure~\ref{fig:RGC05:summary} we examine one ``ON'' cell and one ``OFF'' cell in more depth.  For Neuron 2 we see that the NPGLM approach performs slightly better on the set of data used to fit the model but noticeably better when it comes to generalizing to the repeat stimulus.  This highlights the ability of such a nonparametric approach to generalize well -- even if the data is sparse in the number of spikes as is the case with Neuron 2.  Contrast this to Neuron 4 where while still outperforming the standard approach on the test the difference is not as pronounced.  

\section{Discussion}
In this paper we proposed NPGLM, a nonparametric scheme for making automated inferences for temporal filters that characterizes the influence of external event and stimuli as well as self-generated cognitive decisions and behavior to neural spikes.
NPGLM infers time convolutional filters in autoregressive point process models by utilizing sparse Gaussian process prior and variational inference in a computationally efficient manner.
We verified that NPGLM inference is robust against variations in the sparsity of the GP approximation, and it produces reasonable posterior inference even in the small data regime.
Contrast to previous approaches that model the filters with basis functions, we sidestep the burdensome task of basis function selection and rigorously defining their fixed temporal properties.
We verified the competency of our method on retinal ganglion cell recordings, where it was able to infer filters consistent with previous approaches using basis functions.

While we used binned data in this study, we posit that using spike times at a higher time resolution may reveal structures or features not readily seen using other methods thanks to the infinite time resolution of GPs.  
Further extensions can include tractable methods of inference for determining a more optimal convolution window, pruning inducing points that are not necessary, and adding more inducing points when appropriate.

\section*{Acknowledgement}
This work is supported by NSF CAREER IIS-1845836, IIS-1734910, and NIH UF1-NS115779.

\bibliographystyle{hunsrt}

\newpage
\section*{\hfil Supplement \hfil}
\section*{RGC Analysis Extended}
We continue comparison of inference under NPGLM to the standard basis centric GLM approaches outlined in the paper. Presented is analysis of another set of parasol RGC recordings from the data set previously analyzed in ~\cite{Pillow2005, doi:10.1152/jn.01171.2003}.  This set compromises five ''ON'' cells and four ''OFF'' cells where the stimulus is binary white noise refreshing at a rate of 120 $Hz$ and a contrast of $48\%$. 

Compared to the basis approaches described the model presented is not concave in its parameters -- making a good initialization pertinent.  We initialize the mean of each variational distribution as well as the parameters of the history filter's mean using a GLM that is fitted prior to optimizing the model described.  More specifically,

\begin{align}
    \vm_h^{(0)} &= K_{u_h,u_h} \left(K_{u_h, X_h} K_{X_h,u_h}\right)^{-1}K_{u_s, X_s} \left(\hat{\vm}_h - \vm(X_h)\right) + \vm(u_h) \\
    \vm_s^{(0)} &= K_{u_s,u_s} \left(K_{u_s, X_s} K_{X_s,u_s}\right)^{-1}K_{u_s, X_s} \hat{\vm}_s
\end{align}

Where $\hat{\vm}_h$ and $\hat{\vm}_s$ are the previously fitted means evaluated at the points spanning the initial convolutional window for the history and stimulus filter respectively. Taking advantage of the convexity of the ELBO, $\mathcal{L}$, ~\cite{Arridge_2018} in the variational parameters $\vm$ and $\vS$ with all other parameters fixed their optimal values can also be found before optimizing the hyperparameters and inducing point locations.  Further optimization is performed using a conjugate gradient algorithm which has shown success dealing with non-convex objectives ~\cite{PMID:28065610, latimer_inferring_2019}.  

Fitting of model parameters for NPGLM, GLM-ARD, and GLM-MLE are all performed on 5 seconds of observed stimuli/spiking history with the data binned at a resolution of 1 $ms$. Models are compared with regard to their normalized likelihoods against a homogenuous Poisson process, giving the amount of nats conveyed per spike. This is done for the training data, a 50 $s$ set of holdout data, and 17 trials of exposure to the novel repeat stimulus.  Model parameters (number of basis functions/inducing points) for all three methods of inference remain the same as detailed in the paper. Presented below, in Fig ~\ref{fig:RGC05:summary_supp_1}, are the resultant filters under all three methods of inference and a more in depth comparison between NPGLM and GLM-ARD for each neuron.

\begin{figure}[H]
    \centering
    \includegraphics[width=\textwidth]{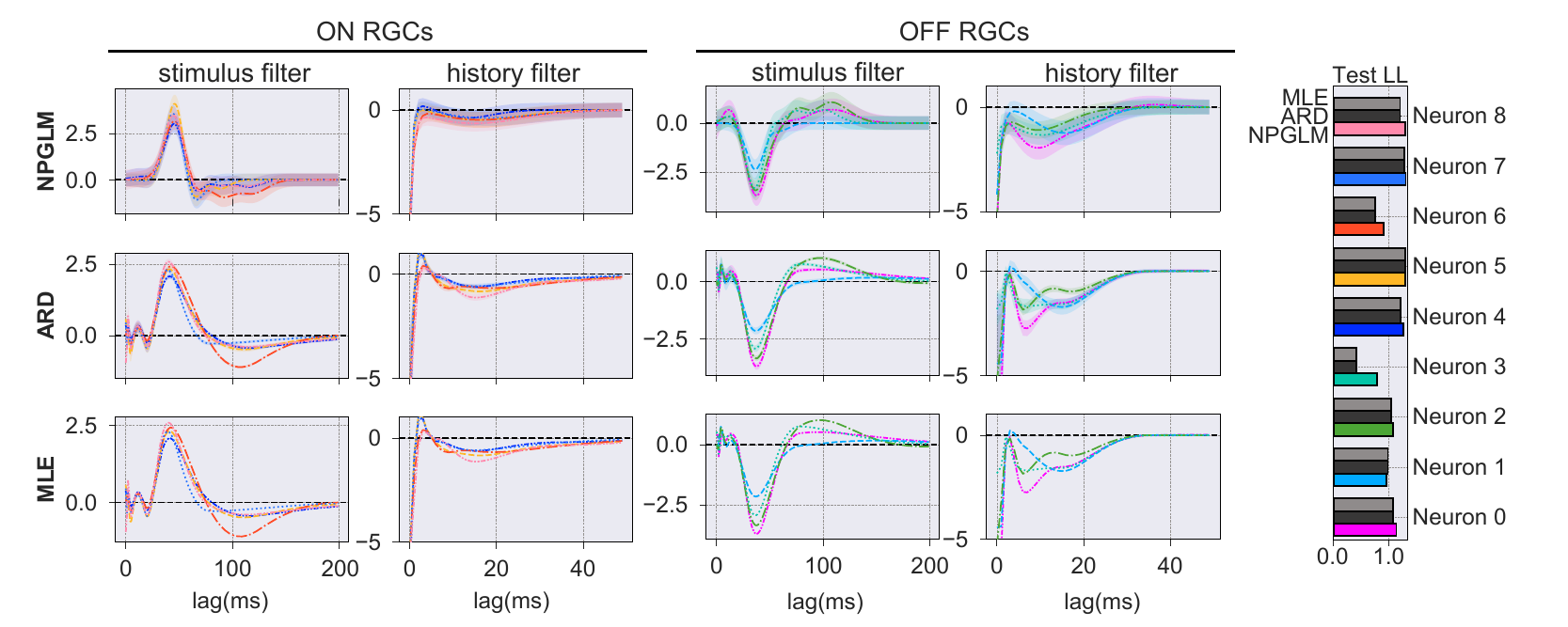}
    \caption{Fits under all three models for 9 RGCs (5 ''ON'', 4 ''OFF''). NPGLM outperforms the bases centric approaches in terms of normalized log-likelihood sans cell 1.
    }
    \label{fig:RGC05:summary_supp_1}
\end{figure}

\newpage

\begin{figure}[H]
    \includegraphics[width=0.9\textwidth]{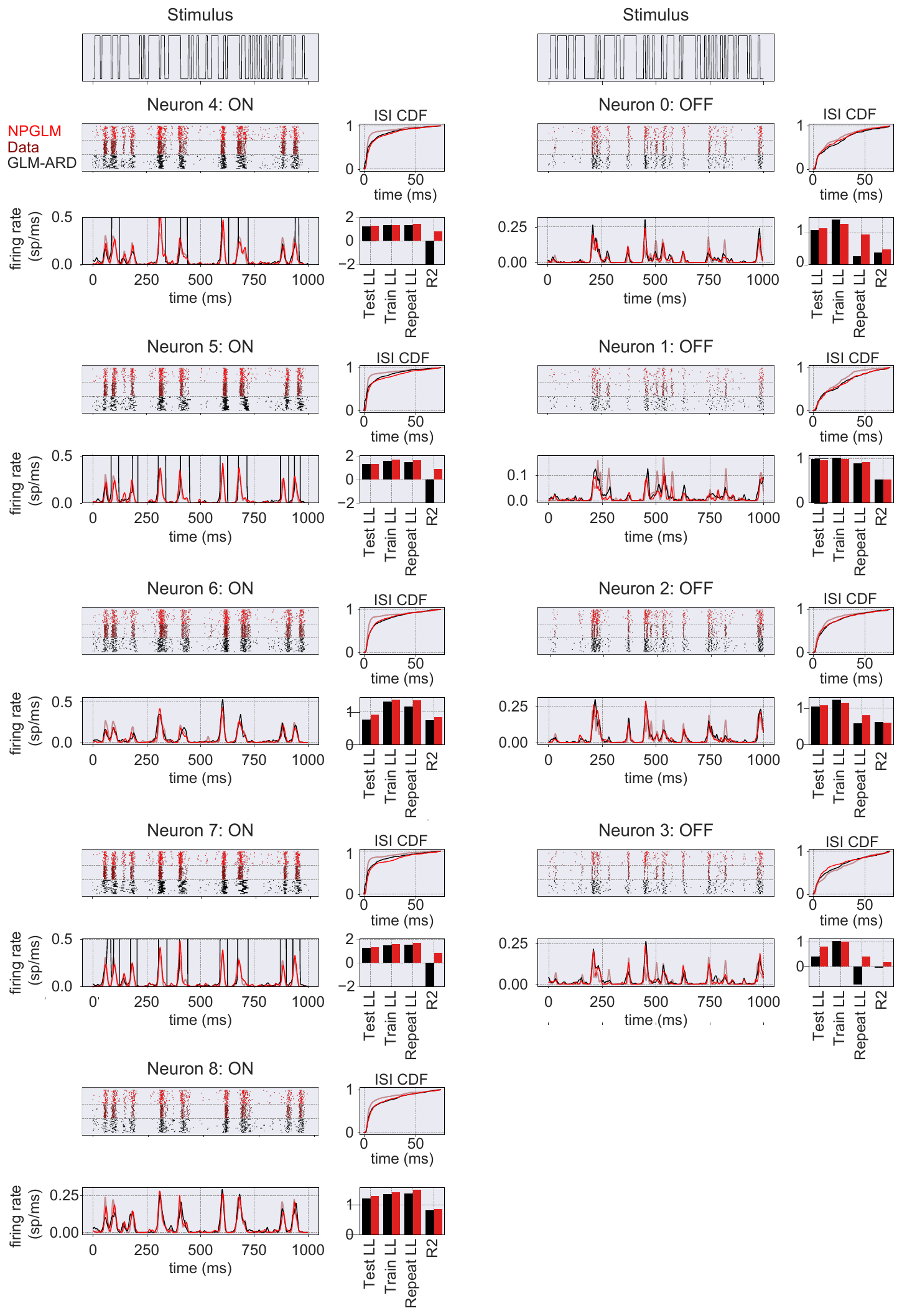}
    \centering

    \caption{\textbf{NPGLM demonstrates the capability to generalize well -- always performing better on the novel repeat stimulus}.
        Generated spike trains, corresponding raster plots, interspike intervals, and four goodness of fit metrics (test/train/repeat set normalized log-likelihoods and R2 w.r.t generated spike trains)
    }
    \label{fig:RGC05:summary_supp_2}
\end{figure}
\newpage

The results presented show that the proposed methodology can \textbf{a)} generalize well to unseen data, and \textbf{b)} generate spike trains that are statistically similar to the underlying data.  We can also see that in cases where the bases centric approach tends to "run away" or vastly overestimate the spiking rate the proposed approach stays in line with the data.

\subsection{Parameter Heuristics}
In general, sans the history filter, we worked with inducing point spacings of 10ms - 15ms, $\alpha=500$, and $\nu=1000$. When only using the DSE kernel alone closer spacing was required to capture the refractory features present in the history filter. Hyperparameters were allowed only to change within 20\% of their value during each new 'round-robin' session of training, which presumably helped to alleviate drastic movements to local minima of the likelihood.  While joint optimization was possible and examined, it was seen to favor one or two filters in particular.  Heuristically, this may have been avoided by introducing an additional loss term to the ELBO however, we did not examine this in too much detail.  Expansion of the design matrix was limited to 50 $ms$ on each update until full coverage of the inducing point range was achieved.

\newpage

\end{document}